\documentclass{article}
\usepackage{ijcai22}

\usepackage{times}
\usepackage{soul}
\usepackage{url}
\usepackage[hidelinks]{hyperref}
\usepackage[utf8]{inputenc}
\usepackage[small]{caption}
\usepackage{subcaption}
\usepackage{graphicx}
\usepackage{amsmath}
\usepackage{amssymb}
\usepackage{amsthm}
\usepackage{booktabs}
\usepackage{algorithm}
\usepackage{algorithmic}
\usepackage{xcolor}
\newcommand{\shuair}[1] 
{
\textbf{\color{blue}{#1}}
}

\title{Vision Transformers: State of the Art and Research Challenges}

\author{
Bo-Kai Ruan$^1$ \and Hong-Han Shuai$^1$ \and Wen-Huang Cheng$^1$\\
\affiliations
$^1$National Yang Ming Chiao Tung University\\
\emails
justin.ee08@nycu.edu.tw,
hhshuai@nycu.edu.tw,
whcheng@nycu.edu.tw
}

\begin{document}

\maketitle

\begin{abstract}
  Transformers have achieved great success in natural language processing. Due to the powerful capability of self-attention mechanism in transformers, researchers develop the vision transformers for a variety of computer vision tasks, such as image recognition, object detection, image segmentation, pose estimation, and 3D reconstruction. This paper presents a comprehensive overview of the literature on different architecture designs and training tricks (including self-supervised learning) for vision transformers. Our goal is to provide a systematic review with the open research opportunities.
\end{abstract}

\section{Introduction}

Transformers \cite{vaswani2017attention} have originally achieved a great success in natural language processing \cite{devlin2018bert,radford2018improving} and can be adopted in various applications, including sentiment classification, machine translation, word prediction, and summarization. The key feature of transformers is the self-attention mechanism, which helps a model learn the global contexts and enables the model to acquire the long-range dependencies. 
Motivated by the great success in natural language processing, transformers have been adopted by computer vision tasks, leading to the development of vision transformers. Vision transformers have become prevalent in these years and have reached considerable success in many fields such as image classification \cite{dosovitskiy2021an,liu2021swin}, video classification \cite{arnab2021vivit}, object detection \cite{carion2020end,fang2021you}, semantic segmentation \cite{ijcai2021-165,ijcai2021-112}, and pose estimation \cite{ijcai2021-188}.

Despite the success of the architecture, there are still several drawbacks which should be addressed, \textit{e.g.}, data-hungry, lack of locality and cross-patch information. Therefore, a recent line of research has been proposed to further enhance vision transformers. This survey introduces important ideas to solve the problems and aims to shed light on these topics for future research. In addition, as self-supervised learning methods play an important role in vision transformers, we also present several self-supervised learning methods employed on vision transformers. The rest of the paper is organized as follows. We start with the preliminary of transformers and vision transformers, and then introduce variant architectures of vision transformers. Afterward, the training tricks used in vision transformers are presented, together with the self-supervised learning. Finally, we conclude this paper and discuss the future research directions and challenges.
\section{Preliminary}
\subsection{Self-attention}
The attention mechanism is one of the most beneficial breakthroughs in deep learning research, which measures the importance of a feature that contributes to the final results. Using an attention mechanism often teaches a model to concentrate on specific features. For self-attention, the input and the output size are the same, while the self-attention mechanism allows the interaction between the inputs and discovers which they should pay more attention to. Afterward, each output is enhanced by the weighted inputs according to the attention scores. For instance, given a sentence stating ``a dog is saved from pond after it fell through the ice,'' self-attention can enhance the embedding of ``it'' by attending to ``dog''. The self-attention is designed to help the model learn the global contexts under the property of long-range dependencies.  

\begin{figure}[t]
    \centering
    \begin{subfigure}[b]{\linewidth}
        \centering
        \includegraphics[width=\linewidth]{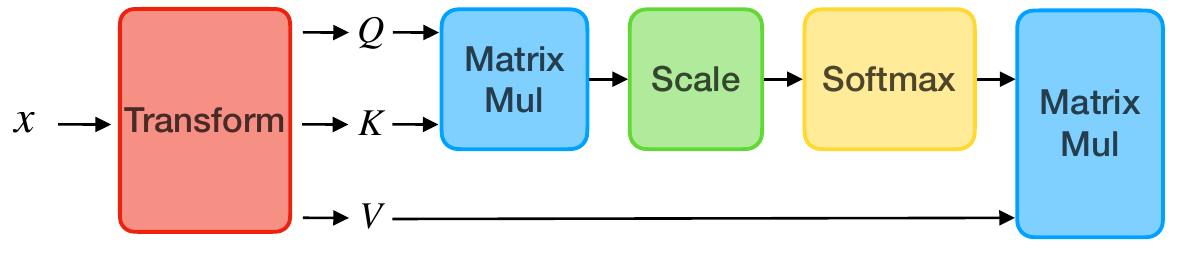}
        \caption{}
        \label{fig:attn}
    \end{subfigure}
    \begin{subfigure}[b]{\linewidth}
        \centering
        \includegraphics[width=\linewidth]{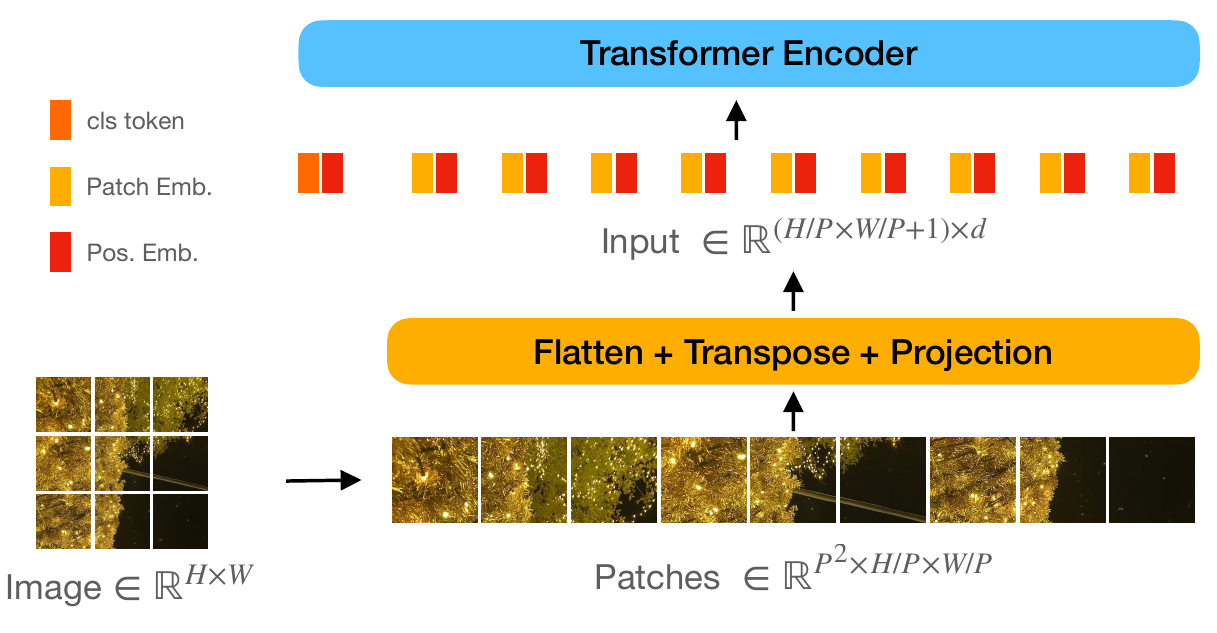}
        \caption{}
        \label{fig:vit}
    \end{subfigure}
    \caption{Illustration of (a) self-attention and (b) image to sequence with $P=3$, $C=1$ and embedding dimension $d$.}
    \label{fig:attn_patch}
\end{figure}

Fig.~\ref{fig:attn} illustrates the self-attention module. Let $X \in \mathbb{R}^{L \times d}$ denote a sequence of vectors $(x_1, x_2,$ $\cdots, x_L)$, where $d$ is the embedding dimension of each vector. To help the model learn the relations between each vector, query, key, and value matrices are projected from $X$ with linear layers and denoted by $Q$, $K$, and $V$, respectively. For instance, the query matrix is obtained by projecting $X$ with a linear layer $W_q$, \textit{i.e.}, $Q=XW_q$. Specifically, the attention weights $E$ are calculated by the normalized product of $Q$ and $K^T$ with the softmax function, \textit{i.e.}, 
\begin{equation}
    E = \text{softmax}(\frac{QK^T}{\sqrt{d}}).
\end{equation}
Softmax function normalizes the attention weights of each query-key pair within $(0, 1)$, where $1$ means the most important, and $0$ means useless information. Finally, the output features $O$ are enhanced by applying the attention weights $E$ to $V$ as follows:
\begin{equation}
    O=\textit{Attention}(Q, K, V) = EV =\text{softmax}(\frac{QK^T}{\sqrt{d}})V.
\end{equation}

\paragraph{Multi-Head Attention.} In order to learn variant representations at different positions, the input $X$ is transformed into $n_h$ different representations (heads), denoted by $h_1,\cdots,h_i,\cdots,h_{n_h}$. The attention is first computed for each head $h_i$ with $Q^i$, $K^i$, and $V^i$ with projection matrices $W_q^i$, $W_k^i$, and $W_v^i$, respectively.
\begin{equation}
    h_i = \textit{Attention}(Q^i, K^i, V^i).
\end{equation}
Afterward, all the heads are concatenated to form the multi-head representations $H$ as follows:
\begin{equation}
    H = \textit{Concat}(h_1, h_2, ..., h_{n_h}).
\end{equation}
Finally, $H$ is transformed back to dimension $d$ with projection matrix $W^O$, \textit{i.e.}, $O=HW^O$. As the number of relations between inputs is usually unknown, multi-head attention is commonly-used to capture different relations between input elements in a data-driven manner.
\subsection{Transformer}
\begin{figure}[t]
    \centering
    \includegraphics[width=0.8\linewidth]{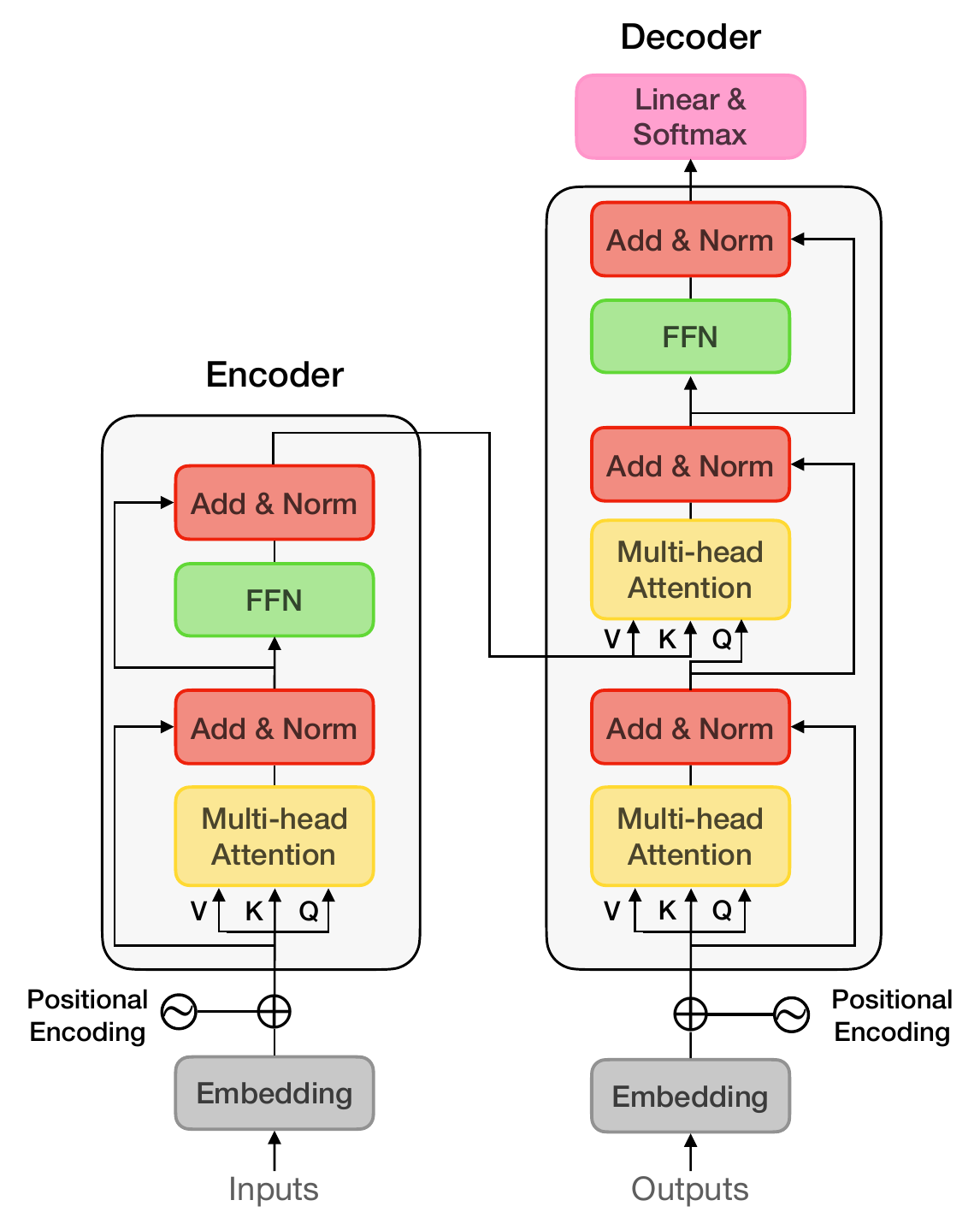}
    \caption{Architecture of Transformer.}
    \label{fig:transformer}
\end{figure}

Transformer, proposed by Vaswani et al. \cite{vaswani2017attention}, are ubiquitous in NLP tasks~\cite{devlin2018bert,brown2020language,ijcai2021-537}. Fig.~\ref{fig:transformer} illustrates the architecture of Transformer. A single transformer block can be separated into \textit{Encoder} and \textit{Decoder}, which both can be further decomposed into 1) self-attention, 2) Position-wise Feed-Forward Networks, and 3) positional encoding.
\paragraph{Position-wise Feed-Forward Networks.} Feed-Forward Networks (FFN) are mainly composed of Fully-Connected layers and can be written in:
\begin{equation}
    \textit{FFN}(X) = \delta(XW_1+b_1)W_2+b_2,
\end{equation}
where $W_1 \in \mathbb{R}^{d \times h}$ and $W_2 \in \mathbb{R}^{h \times d}$ are the weight and $b_1$ and $b_2$ are the bias terms. $\delta$ is ReLU activation function. $h$ is the hidden dimension and is usually set to $4d$.
\paragraph{Positional Encoding.} Since the self-attention relaxes the sequential order, transformers use positional encoding to maintain the ordinal information. The author proposes to use sine and cosine functions with different frequencies to derive the positional encoding, which allows the model to learn the relative positions since any positional encoding can be linearly combined by other positional encodings. Specifically, the $i$-th dimension of the positional encoding at position $pos$, denoted by $\textit{PE}_{(pos, i)}$ can be calculated as follows:
\begin{equation}
\textit{PE}_{(pos, i)} = 
\begin{cases}
    \sin(pos/10000^{i/d}) & \text{if }i \text{ is even,} \\
    \cos(pos/10000^{(i - 1)/d}) & \text{otherwise,}
\end{cases}
\end{equation}
where $d$ is the dimension of the input features.
\paragraph{Encoder.} A single encoder block has two sub-layers, \textit{i.e.}, multi-head attention and position-wise feed-forward networks. Each sub-layer is followed by residual connection \cite{he2016deep} and layer normalization \cite{ba2016layer}. The outputs of the encoder will be sent into the decoder as $K, V$ in the second multi-head attention layer.
\paragraph{Decoder.} Each decoder block is made by three sub-layers. The first two layers are multi-head attention blocks, and the final one is position-wise feed-forward networks. Similar to encoder block, all three layers are followed by residual connection and layer normalization as well. Since the decoder is auto-regressive, \textit{i.e.}, sequentially predicting a new result only based on previous predictions, the multi-head attention layers in the decoder utilize the mask operation to only attend the predicted results for preventing the violation of causality.

\subsection{Vision Transformer (ViT)}
ViT \cite{dosovitskiy2021an} replaces all the CNN structures with several transformer layers and reaches state of the art performance on image recognition and are known as the pioneer of vision transformers. ViT contains three segments: 1) patch and positional embedding, 2) Transformer encoder, and 3) multi-layer perceptron (MLP) head.
\paragraph{Patch and Positional Embedding.} In order to transform an image $X \in \mathbb{R}^{C \times H \times W}$ into a 1D sequence of vectors, an image will be transformed into $X_p \in \mathbb{R}^{N \times (P^2C)}$ as illustrated in Fig.~\ref{fig:vit}, where $P\times P$ is the resolution of each patch, and the number of patches is $N = HW/P^2$. The patches are then projected into patch embeddings by a linear layer. Akin to BERT \cite{devlin2018bert}, a learnable class token $x_{cls} \in \mathbb{R}^{1 \times (P^2C)}$ is added to $X_p$. Afterward, the positional embeddings are added to the output $[x_{cls}; X_p]$ to obtain the positional information.
\paragraph{Transformer Encoder.} Instead of using both encoder and decoder, ViT only utilizes the Transformer encoder since the goal is to find a better representation rather than autoregressive prediction. The image patches are transformed into sequences and then sent into the encoder. The only difference is that the Layer Normalization is added before the sub-layers \textbf{(pre-norm)}~\cite{xiong2020layer}.
\paragraph{Class Tokens.} The class token is trained to include the class information and can be used to represent the entire features. Therefore, it can be used to classify the data into different categories.
\paragraph{MLP Head.} MLP head is commonly-used for the downstream tasks. Specifically, let $Z=[Z_0,Z_1,\cdots,Z_N]$ denotes the output of the Transformer encoder. As the class token prepended to the input sequence can be regarded as the image representation, the final result $y$ is determined based on the first token of the output $Z_0$ from encoders through a single MLP layer, \textit{i.e.},
\begin{align}
    Z &= \text{TransEncoder}([x_{cls}; X_p]), \\
    y &= \text{MLP}(Z_0).
\end{align}


\section{Variant Architectures}
Despite the promising performance of ViT, it still suffers from several issues. For instance, ViT requires to be trained on a large dataset. As ViT is initially trained on a JFT dataset \cite{sun2017revisiting} with $\approx 300M$ images before being fine-tuned on ImageNet ($\approx 1.2M$ images) \cite{ILSVRC15}. If the dataset is insufficient, the model may perform worse than CNN-based approaches. Although the pre-trained weights can be used on various tasks, many datasets are not transferable or even with an inferior performance. Moreover, ViT is not a general-purpose backbone since it is only suitable for image classification but not for the dense prediction, such as object detection and image segmentation, due to the patch partitions. 

\begin{figure}[t!]
    \centering
    \includegraphics[width=\linewidth]{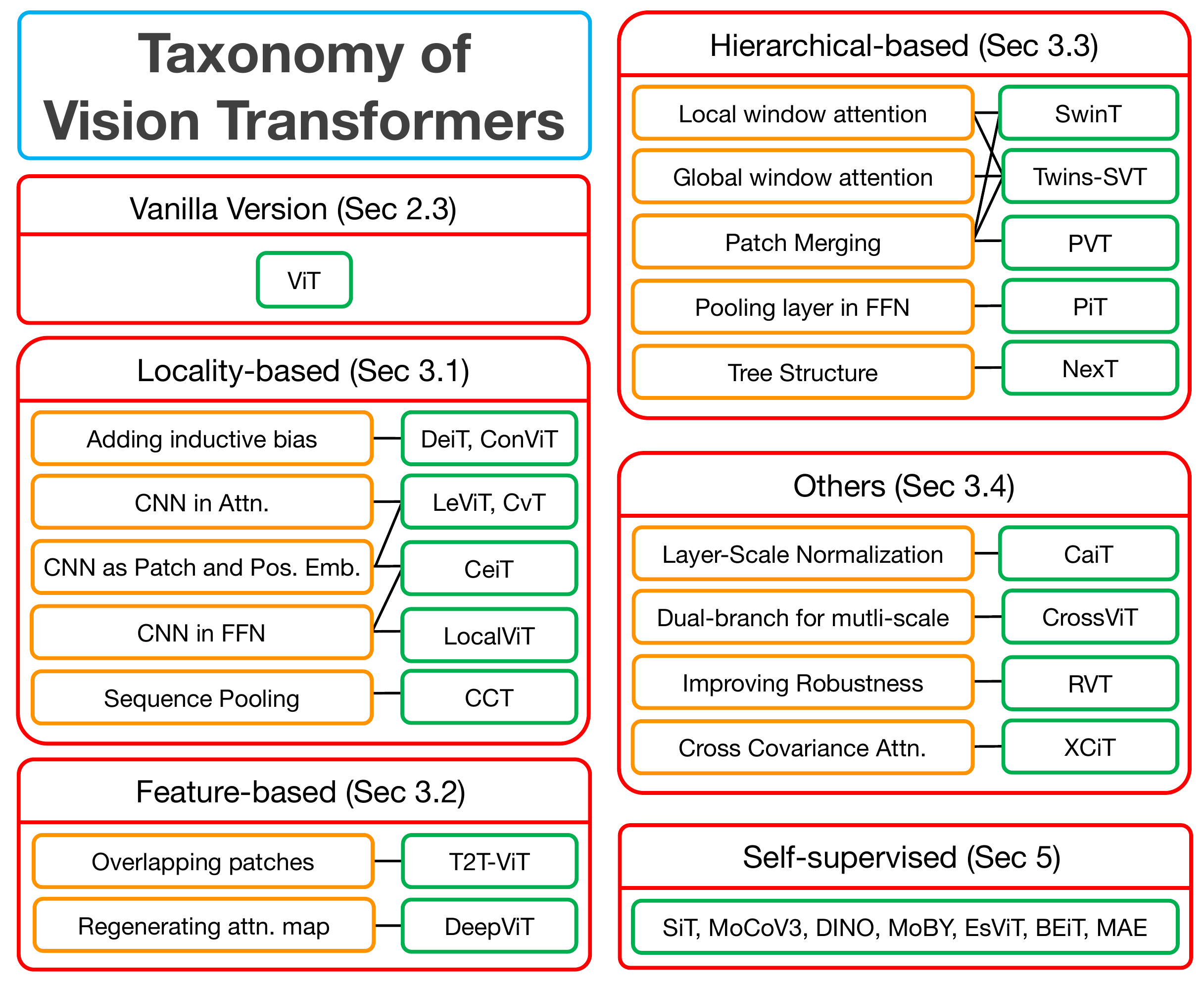}
    \caption{Taxonomy of Vision Transformers.}
    \label{fig:classification}
\end{figure}

Fig.~\ref{fig:classification} shows the taxonomy of vision transformers with three mainstream directions. Specifically, Sec.~\ref{sec:local} first introduces the \textit{locality-based models}, which manage to add the locality into the architectures. Next, \textit{feature-based models} are introduced in Sec.~\ref{sec:feature}, which aim to diversify the feature representations. Finally, \textit{hierarchical-based models} are presented in Sec.~\ref{sec:hierarchical}, which reduce the feature size layer-by-layer to increase the inference speed. Some architectures that are not classified into the above categories are included in Sec.~\ref{sec:other}. Please note that these models are put into certain categories, but these categories are not mutual-exclusive.

\subsection{Locality-based Models}\label{sec:local}
ViT, which lacks locality and translation equivalence, usually performs worse than CNN. Therefore, researchers start to include the CNN structures into the vision transformer since the convolution kernels help the model capture the local information. As such, adding locality from CNN improves the data efficiency of vision transformers, resulting in a better performance on a small dataset. In the following, we introduce several approaches considering the locality.
\paragraph{DeiT} \cite{touvron2021training} uses a CNN as a teacher model to train a vision transformer, which utilizes knowledge distillation \cite{hinton2015distilling} to transfer the inductive bias to a vision transformer and applies stronger data augmentation for input data. This approach allows others to train a vision transformer from scratch without the requirement of pre-training on a large dataset.
\paragraph{ConViT} \cite{d2021convit} is similar to DeiT as ConViT also combines the inductive bias of CNN into models. Instead of using knowledge distillation, ConViT includes \textit{Gated Positional Self-attention (GPSA}), which can be initialized as a convolutional layer \cite{Cordonnier2020On} for capturing the local information at the beginning of the training stage. As such, ConViT can utilize the advantages of soft inductive bias of CNN without being limited to CNN. In other words, GPSA allows vision transformers to be the same as CNN to improve the data efficiency on small datasets but can be better than CNN when there are infinite data used in training. 
\paragraph{LeViT} \cite{graham2021levit} gets the embeddings from four convolutional layers, which the model can extract the local features at the beginning and also reduce the input size. On top of that, LeViT further lowers the input size in some attention blocks, which speeds up the inference time. These advantages help strike the balance between accuracy, data efficiency and the training speed.
\paragraph{CeiT} \cite{yuan2021incorporating} obtains the embedding features directly from convolutional blocks to add the locality in the beginning of the model, similar to LeViT. Moreover, the authors include a depth-wise convolutional layer in FFN to encourage the model to extract the local features. In order to exchange the class information in different layers, CeiT utilizes \textit{Layer-wise Class-token Attention} to collect different class representations by computing the self-attention on class tokens.
\paragraph{CvT} \cite{yuan2021incorporating} also obtains the embeddings with convolutional layers. Besides, CvT uses convolutional layers to create query, key, and value before self-attention operation, which provides the local spatial information. Furthermore, CvT also shrinks the size of the key and the value by using $stride=2$ to accelerate the computation.
\paragraph{LocalViT} \cite{li2021localvit} is designed to use convolutional layers in FFN to extract the local features in every transformer blocks. The authors also try to apply different activation functions (ReLU6, h-swish) and architectures (SE-Block, ECA module) in FFN layers to improve the performance.
\paragraph{CCT} \cite{hassani2021escaping} is proposed to solve the data-hungry problem to make vision transformers perform well on small datasets. Specifically, CCT gets the embeddings with convolutional layers and makes the input shape flexible by removing the positional embeddings. On top of that, CCT incorporates the \textit{Sequence Pooling} at the end of the transformer layers to compute the weights of the output sequence.


\subsection{Featured-based Models}\label{sec:feature}
These models put their efforts on diversifying the features, \textit{e.g.,} token maps, attention maps in vision transformers. Having distinct feature maps indicates that the model can extract various features, which allows the model to perform well.
\paragraph{DeepViT}\cite{zhou2021deepvit} is created after the authors examine the attention maps of ViT and find out that the attention collapsing takes place in the deeper layer. This problem hinders the model to be representative and would lower the performance. The authors alter the self-attention layers and provide a learnable transformation matrix after the attention layer to address the issue by stimulating the model to generate a new set of attention maps.
\paragraph{T2T-ViT}\cite{yuan2021tokens} is designed after the authors observe the feature maps reshaped from the tokens and point out that most of the token maps are meaningless in ViT. To improve the diversity of the token maps, the authors inserts a T2T-module in the beginning of the model. T2T-module is made up of few T2T-Transformers. These T2T-Transformers can be viewed as the original transformer layers or can be replaced by the Performer used in \cite{choromanski2021rethinking}. Between the T2T-Transformers is the T2T-process, which crops the images into several overlapping patches. The overlapping strategy enables the model to share the information between the neighbors to improve the features diversity.

\subsection{Hierarchical-based Models}\label{sec:hierarchical}
The original version of ViT is notorious for its heavy computation. Significantly, the cost raises when the input size increases. Therefore, decreasing the feature size would definitely help reduce the training time. Below are some approaches to solve this issue.
\paragraph{PVT} \cite{wang2021pyramid} is mainly designed for solving dense prediction problems (e.g., object detection and semantic segmentation). It uses \textit{Spatial Reduction Layer} to reduce the computation by decreasing the dimension of $K$ and $V$. The feature size is reduced by a patch embedding layer at the beginning of each stage. Due to its pyramid structure, the model can generate multi-scale feature maps and can be trained faster with its smaller feature size.
\paragraph{PiT} \cite{heo2021rethinking} includes \textit{Pooling Layer}, which uses a depth-wise convolutional layer to achieve the dimension reduction. The authors also test PiT on different robustness benchmarks and get outstanding performance.
\paragraph{Swin-Transformer} \cite{liu2021swin} is proposed to derive a general-purpose backbone based on ViT, which can be used for different applications, \textit{e.g.}, image classification and semantic segmentation. Since applying self-attention pixel-by-pixel results in a tremendous computation complexity, Swin-Transformer forms a hierarchical structure, which merges patches after each Swin-Transformer block and therefore has approximate linear computation time complexity to input image size due to the computation of self-attention only within each local shifted window.
\paragraph{Twins-SVT} \cite{chu2021twins} also computes the attention within a shifted window, while the global attention is evaluated after the local window attention. This helps each window retains the outside-window information, similar to the overlapping strategy. Likewise, the patches are merged at the beginning of the stages to form the hierarchical shape.
\paragraph{NexT} \cite{zhang2021aggregating} is created on a different strategy to perform the hierarchical computation. Specifically, an image is first partitioned into $n$ blocks and every four blocks are merged after a transformer layer. Moreover, \textit{Gradient-based Class-aware Tree-traversal} is proposed to visualize the most critical path from child to root, which reveals how the model makes the decision given an input image. It is worth noting that NexT can be used to generate images by turning the tree upside down due to its tree structure.


\subsection{Others}\label{sec:other}
In the following, we introduce several promising directions for improving ViT which are not classified before.
\paragraph{CrossViT} \cite{chen2021crossvit} can extract multi-scale features from two branches. These branches are L-Branch and S-Branch, where the former uses a larger patch size and the latter uses a smaller patch size. By using the dual-branch structure, the model is able to obtain different scales of spatial information. To fuse the spatial information between the branches, \textit{Cross Attention} is applied by inserting the class token from one to the other. 
\paragraph{RVT} \cite{mao2021towards} is designed after studying different components in ViT with regards to the robustness, \textit{e.g.}, accuracy under adversarial attacks. Based on the results, RVT decides to 1) remove the class token, which is not important to a vision transformer, by averaging the features and 2) add CNN to the embedding and the FFN layers for increasing locality and 3) use more attention heads for obtaining different features. Moreover, the original self-attention is replaced by \textit{Position-Aware Attention Scaling (PAAS)}, which adds a learnable matrix to the self-attention for showing the importance of each $Q$\texttt{-}$K$ pair. PAAS is proved to suppress the unrelated signal in the noise input. Moreover, \textit{Patch-wise Augmentation} applies different augmentations on different patches to diversify the training data.
\paragraph{CaiT} \cite{touvron2021going} employs the normalization with \textit{LayerScale}, which uses learnable factors to adaptively normalize the features. The normalization speeds up the converge rate and allows the deep model to be trained well. CaiT also includes \textit{Class Attention Layer} for computing the attention between the class embedding and the overall features to have better knowledge on the inputs.
\paragraph{XCiT} \cite{ali2021xcit} is proposed to reduce the time complexity of the self-attention. The reduction is done by \textit{Cross-Covariance Attention}, which uses the transposed version of self-attention, \textit{i.e.}, the self-attention is computed on the feature channels but not on the tokens. As such, the time complexity is reduced from $O(N^2d)$ to $O(N^2d/h)$. Moreover, \textit{Local Patch Interaction Block} is employed to use depth-wise convolutional networks to further extract the information between different patches.

\section{Training Tricks for Vision Transformer}
To better train a vision transformer, several tricks are proposed to increase the diversity of the data and to improve the generality of the model.
\paragraph{Data Augmentation} is used to increase the diversity of training data, \textit{e.g.}, translation, cropping, which help a model learn the main features by altering the input patterns. To find out the best combination for a variety of datasets, AutoAugment \cite{Cubuk_2019_CVPR} and RandAugment \cite{cubuk2020randaugment} are designed to search for a better combination. These augmentation strategies are proved to be transferable to different datasets. 
\paragraph{Exponential moving average (EMA)} is often added to stabilize the training process. Let $\theta^l$ and $\theta^{l'}$ respectively denote the model parameters in the $l$-th iteration and the parameters updated by an optimizer. The model parameters in the $(l+1)$-th iteration can be calculated by
\begin{equation}
    \theta^{l+1} = \lambda \theta^l + (1 - \lambda)\theta^{l'},
\end{equation}
where $\lambda$ is a hyperparameter in range $[0, 1]$. EMA stabilizes the training process by smoothing old and new parameters.
\paragraph{Stochastic Depth (SD)} \cite{huang2016deep} is first proposed to train deep networks such as ResNet \cite{he2016deep}, which drops the entire block as a regularization method. Similarly, when training a vision transformer, stochastic depth randomly leaves some samples to be 0 after an attention or a FFN layer but before the residual connection. This step can be viewed as randomly replacing the networks by identity function for some samples.
\paragraph{Fixing Resolution Discrepancy} \cite{touvron2019fixing} is to relax the discrepancy between training and testing image size resulted from random cropping in data augmentation. The authors find that upsampling the training data can mitigate the discrepancy caused by different resolutions. This is why many models are trained with a size of $384$ or bigger.

\section{Self-supervised Learning in Vision Transformer}\label{sec:ssl}
Self-Supervised Learning (SSL) trains a model as supervised learning but only uses data itself to create labels instead of manual annotation. SSL has a considerable advantage over exploiting the data, especially useful for those extensive datasets. It also helps the model learn essential information lies in data and makes the model robust and transferable. 

Recently, SSL in computer vision can be mainly categorized into pretext tasks and contrastive learning. The former is to design a particular job for a model to learn before fine-tuning on downstream tasks, \textit{e.g.}, predicting the rotation degree, coloring, or solving the jigsaw puzzles. In contrast, the latter generates similar features for the same-class data and pushes away other negative samples. In the following, we introduce several SSL methods used in the vision transformer.
\paragraph{SiT} \cite{atito2021sit} includes two pretext tasks with the contrastive loss after passing input data through a vision transformer. These tasks include predicting the rotation degree of $(0°, 90°, 180°, 270°)$ and the image reconstruction. The image is first augmented and partitioned into different patches. The rotation task is to match the predicting degree and the actual degree of the input image. The reconstruction task is to reconstruct the augmented image back to the original image. Finally, the contrastive loss maximizes the similarity between the same inputs.
\paragraph{MoCoV3} \cite{chen2021empirical} is designed explicitly for vision transformer based on V1 and V2 \cite{he2020momentum,chen2020improved}. Each image has two different versions generated by data augmentation and the images of different versions are fed into two different encoders. The model then learns to reduce the difference between the two identical images and increases the distance of those negative samples, where the difference is measured by \textit{InfoNCE} \cite{oord2018representation}. 
Other modifications are 1) removing the \textit{memory queue}, which is able to reduce the requirement of a substantial batch size, and 2) adopting the symmetrized loss. Suppose the first encoder outputs $q_1, q_2$ and the second encoder outputs $k_1, k_2$, the symmetrized loss $\mathcal{L}$ is computed by:
\begin{equation}
    \mathcal{L} = \textit{infoNCE}(q_1, k_2) + \textit{infoNCE}(q_2, k_1).    
\end{equation}
\paragraph{DINO} \cite{caron2021emerging} is also trained with \textit{InfoNCE} but with different methods. The inputs are cropped into global and local views, where the global views have higher resolution. The teacher model can only see the global views while the student model can utilize all the views. During the updating stage, the student model is updated by an optimizer, while the teacher model updates the parameters using EMA with the student model. To avoid collapsing, DINO adds centers, which can be viewed as a bias term, to each teacher output for helping the model generate the uniform distribution. The centers are updated smoothly by the average of the teacher outputs.
\paragraph{MoBY} \cite{xie2021self} incorporates the training strategies from MoCoV2 \cite{chen2020improved} and BYOL \cite{grill2020bootstrap}. Instead of using vanilla ViT as the backbone, MoBY replaces it with Swin-Transformer \cite{liu2021swin}. Additionally, the updating strategy is similar to DINO. On top of that, to avoid a large batch size, MoBY follows MoCoV2 to create \textit{memory queue} for reusing the past features.
\paragraph{EsViT} \cite{li2022efficient} uses a hierarchical transformer for reducing the computational cost. Instead of using positional embeddings, EsViT uses the \textit{Relative Position Bias} to avoid the positional information being affected by the different cropping resolutions. Furthermore, EsViT uses the same updating method adopted by DINO and MoBY. On top of that, EsViT adds extra regional-level tasks to attach the inter-region relationships.
\paragraph{BEiT} \cite{bao2022beit} is designed based on BERT. The images are separated into patches, which are tokenized into different discrete values by training a discrete VAE \cite{ramesh2021zero}. Afterward, the vision transformer are made to predict the tokens of the masked patches. The ground truth of the tokens are generated with the non-mask patches by a discrete VAE trained in the first stage.
\paragraph{MAE} \cite{he2021masked} is created based on method adopted by autoencoder to train a vision transformer. Distinctly, the input image is first masked by up to $75\%$. These masking strategies applied on images include random, block-wise, and grid. Then, the masked images are encoded into features by a ViT and the decoder decodes the features by another ViT into the original non-mask images. Astonishingly, this strategy ends up with an unprecedentedly high accuracy and the masked images can be well reconstructed.

\section{Challenges and Discussions}
Although existing works have made a considerable success, there are still lots of challenges left to be resolved. In the following, several open challenges and future directions of vision transformers are discussed.
\paragraph{Universal pre-trained weights.} Vision transformers are inspired by the self-attention in Transformer, which is originally used for solving NLP problems. However, the adaption from text to vision is not completely explored. One promising direction is to find out the universal pre-trained weights that are suitable for different kinds of inputs, \textit{e.g.}, texts, images or audio. Currently, \cite{li2021towards} discuss the relationship between texts and visions. More explorations on general conditions enable us to unveil the mask of the underlying principles of transformers.
\paragraph{Fixed input size.} Although self-attention accepts varied sequence lengths, positional embeddings require the fixed length for each input. One approach to deal with this issue is by interpolation, which help expand or compress the embeddings with a given size. Nevertheless, this would cause the information loss when the input size is extremely different from the training size. Currently, the only feasible approach is to directly extract the features from convolutional layers without adding positional embeddings. However, using only the convolutional layers lacks the global positional information.
\paragraph{Robustness.} It is important to test the robustness of a model when the input image are altered or corrupted by some uncontrolled reasons. These alternations include brightness, background, blur, noise, digital artifacts, or even adversarial attack. Despite that RVT examines the robustness of different components, it only focuses on the vision transformer architecture. Other aspects such as data augmentation or learning objective are not explored yet. The former includes what kinds of combination of data augmentation can resist the noise inputs and the latter are associated with the learning criterion that enables the model to filter out the noise attack.
\paragraph{Lightweight model for mobile devices.} Since deep learning has gradually become popular in these years, more and more manufacturers transplant the deep learning models into mobile devices. However, due to the limitations of the size and the cost, the computing resources within a mobile device are not suitable for running a vision transformer. Therefore, reducing the model size is also an important topic for extended usage on mobile devices. Currently, there are only few publications \cite{mehta2022mobilevit} focus on solving the related issue.
\paragraph{Feature Collapsing.} The ability to extract the features highly influence the model performance. As described in Sec.~\ref{sec:feature}, the original vision transformer is subjected to feature collapsing and cannot generate various representations, which is a serious issue when training a deep vision transformer. Current methods often add additional modules (T2T-ViT) or learnable parameters (DeepViT) to solve the problem, which can cause overhead in inference time. Investigation on other methods such as altering the architectures or training strategies, adding additional criterion and adopting different data augmentations are also promising directions for preventing the vision transformers from collapsing.

\section{Conclusion}
We present several vision transformer models and highlight the innovative components. Specifically, variant architectures are introduced to deal with weaknesses such as data-hungry, low efficiency, and weak robustness. These ideas include transferring the inductive bias from CNN, adding locality, strong data augmentation, cross-window information exchange, and reducing the computational cost. We also review the training tricks, as well as self-supervised learning, which trains datasets without requiring any labels but can even reach a higher accuracy than that of the supervised methods. At the end of our paper, we discuss some open challenges for the future research.
    
\bibliographystyle{named}
\bibliography{refer}

\end{document}